\def\year{2021}\relax
\documentclass[letterpaper]{article} 
\usepackage{aaai21}  
\usepackage{times}  
\usepackage{helvet} 
\usepackage{courier}  
\usepackage[hyphens]{url}  
\usepackage{graphicx} 
\usepackage{natbib}  
\usepackage{caption} 
\usepackage{xcolor}
\usepackage{amsmath}
\usepackage{amsfonts}       
\usepackage{multirow}
\usepackage{array}
\usepackage{soul}
\newcolumntype{P}[1]{>{\centering\arraybackslash}p{#1}}

\title{Towards Sample-efficient Apprenticeship Learning from Suboptimal Demonstration}
\author {
    Letian Chen,
    Rohan Paleja,
    Matthew Gombolay\\
}
\affiliations {
    College of Computing, Georgia Institute of Technology \\
    letian.chen@gatech.edu, rohan.paleja@gatech.edu, matthew.gombolay@cc.gatech.edu
}
\begin{document}

\maketitle

\begin{abstract}
Learning from Demonstration (LfD) seeks to democratize robotics by enabling non-roboticist end-users to teach robots to perform novel tasks by providing demonstrations. However, as demonstrators are typically non-experts, modern LfD techniques are unable to produce policies much better than the suboptimal demonstration. A previously-proposed framework, SSRR, has shown success in learning from suboptimal demonstration but relies on noise-injected trajectories to infer an idealized reward function. A random approach such as noise-injection to generate trajectories has two key drawbacks: 1) Performance degradation could be random depending on whether the noise is applied to vital states and 2) Noise-injection generated trajectories may have limited suboptimality and therefore will not accurately represent the whole scope of suboptimality. We present \textit{Systematic} Self-Supervised Reward Regression, S3RR, to investigate systematic alternatives for trajectory degradation. 
We carry out empirical evaluations and find S3RR can learn comparable or better reward correlation with ground-truth against a state-of-the-art learning from suboptimal demonstration framework. 
\end{abstract}

\section{Introduction}
The field of Learning from Demonstration (LfD) seeks to democratize access to robotic assistance by empowering end-users to teach robots by demonstrations, avoiding the need for an army of engineers to manually program and test robots for every task~\citep{seraj2020coordinated}. Due to the democratized access via LfD, applications ranging from healthcare~\cite{gombolay2018robotic}, manufacturing~\citep{wangheterogeneous}, auto-pilot~\citep{pan2017agile}, to assistive jobs~\citep{8624330} have been able to take advantage of expert users in robot learning. However, demonstrations from real end-users are typically suboptimal due to limited cognitive abilities~\cite{newell1972human,nikolaidis2017game}, inhibiting robot learning frameworks from using direct imitation learning \cite{chernova2014robot, ross2011reduction, Paleja2019InterpretableAP}, which learns a direct mapping from a state to the demonstrated action, and therefore only imitates the behavior rather than discovering the latent objective of the demonstrator. 

Inverse Reinforcement Learning (IRL)~\citep{ng2000algorithms,Chen2020JointGA} on the other hand, infers the demonstrator's underlying objective. With the objective represented as a reward function, Reinforcement Learning (RL) can find a policy accomplishing the desired goal. However, similar to imitation learning, classic IRL approaches
~\citep{abbeel2004apprenticeship,Gombolay:2016a} rely on the 
assumption that the demonstration provided is optimal. Maximum-entropy IRL~\citep{ziebart2008maximum} and Bayesian IRL~\citep{ramachandran2007bayesian} relax 
this to stochastic optimality, but in general, it cannot produce a much better policy than the suboptimal demonstrations, limiting the applicability of LfD with na\"ive users. 

Drawing inspiration from Preference-based Reinforcement Learning~\cite{wirth2017survey}, D-REX~ \cite{brown2020better} and SSRR~\cite{Chen2020JointGA} have developed successful learning from suboptimal demonstration frameworks that utilize a performance degradation across noisy trajectories to recover 
the demonstrator's latent objective. SSRR~\cite{Chen2020JointGA} set the state-of-the-art by first utilizing self-supervision to create new trajectories via noise-injection while simultaneously estimating the reward for each trajectory. Next, under the assumption that increased noise will further degrade the trajectory performance, a noise-performance relationship is developed across the generated trajectories, which is then utilized to learn an idealized reward function. Following the learning of a reward function, policy learning occurs via RL to generate a policy that can outperform the demonstrations gathered from suboptimal end-users. 

In this paper, we investigate \textit{systematic} alternatives to self-supervised trajectory generation via noise-injection. Utilizing noise-injection to generate trajectories has two main drawbacks: 1) Noise injection is only effective if the noise is applied in vital states among the trajectory randomly. For example, if the robot is idle for the first two seconds of the demonstration, adding noise during this period will have a very small impact compared with adding noise when the robot is executing a maneuver; 2) Degraded trajectories via noise-injection have increased randomness but may not be able to accurately represent the whole scope of suboptimality. An adversarial/highly sub-optimal policy may perform much worse than a random policy. For example, if the objective is to make a robot cheetah move forward, the expected performance a random policy could achieve is to not move. However, a policy moving backward yields even lower reward, representing an even further degraded instance in the degradation-performance relationship. 
We extend our prior work Self-Supervised Reward Regression (SSRR) \cite{Chen2020LearningFS}, to propose \textit{S3RR, Systematic Self-Supervised Reward Regression} that assesses alternative approaches to generate performance-degraded trajectories. 
Our proposed trajectory generation approaches allows S3RR to recover more accurate reward functions while avoiding the pitfalls of trajectory generation via noise-injection.
We then present empirical results across two simulated robot control tasks and find S3RR can achieve higher reward correlation with ground-truth than SSRR. 
\begin{figure*}[t]
  \centering
  \includegraphics[width = 0.9\linewidth]{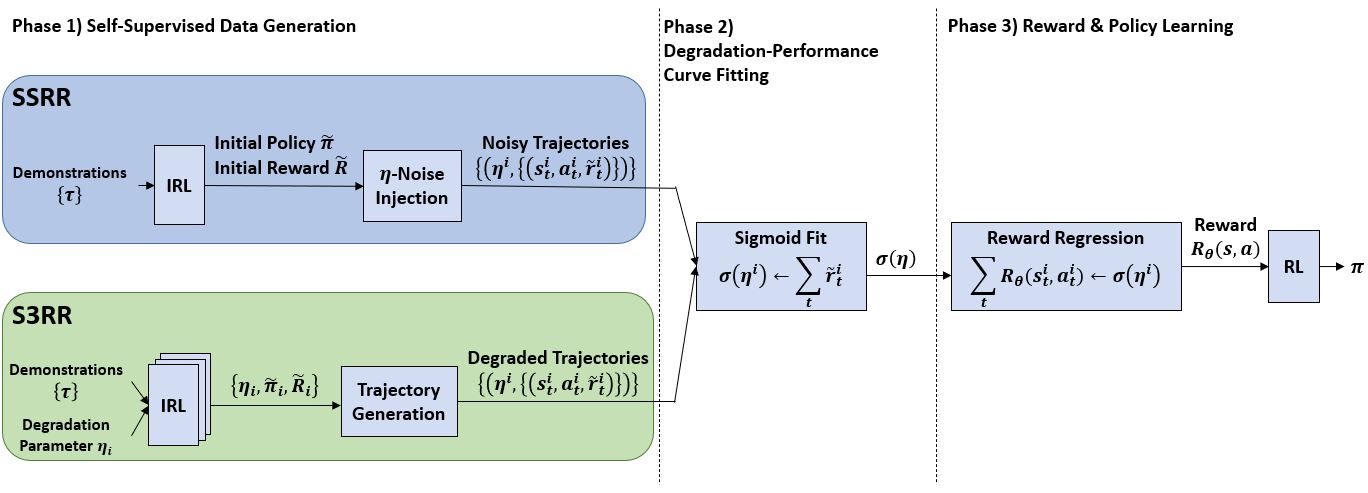}
  \vspace{-12pt}
  \caption{This figure depicts comparison between SSRR and S3RR algorithm pipeline. (Top) SSRR (Bottom) S3RR.}
  \label{fig:ssrr_diagram}
\end{figure*}

\section{Preliminaries}
\label{section:prelim}

In this section, we introduce Markov Decision Processes, IRL, and SSRR~\cite{Chen2020LearningFS}. 
\paragraph{Markov Decision Process --} A Markov Decision Process (MDP)~\citep{white1993survey} $M$ is defined as a 6-tuple $\langle S,A,R,T,\gamma,\rho_0\rangle$. $S$ and $A$ are the state and action spaces, respectively: $s\in S, a\in A$. $R(s,a)$ defines the reward given to an agent for executing action, $a$, in state, $s$. $T(s^\prime|s,a)$ denotes the probability of transitioning from state, $s$, to state, $s'$, when applying action, $a$. $\gamma\in [0,1]$ is the temporal discount factor. $\rho_0(s)$ is the distribution of initial state. A policy, $\pi(a|s)$, defines the probability of the agent taking action, $a$, in state, $s$. The objective for RL is to find the optimal policy, 
$\pi^*=\arg\max_\pi \mathbb{E}_{\tau\sim\pi}\left[\sum_{t=0}^T{\gamma^tR(s_t,a_t)}\right]$, which maximizes cumulative discounted reward in the sampled trajectories, where a trajectory $\tau=\langle s_0,a_0,r_0,\cdots,s_T,a_T,r_T\rangle$ is a sequence generated by the MDP and the policy: $s_0\sim\rho_0(\cdot), a_t\sim\pi(\cdot|s_t), s_{t+1}\sim T(\cdot|s_t,a_t), r_{t}\sim R(s_t,a_t),\ \forall t\geq 0$. We consider the maximum entropy version of RL goal~\citep{ziebart2010modeling}, $\pi^*=\arg\max_\pi\mathbb{E}_{\tau\sim \pi}[\sum_{t=0}^T{\gamma^tR(s_t,a_t)}+\alpha H(\pi(\cdot | s_t))]$, which adds an entropy bonus (i.e., $H(\cdot)$) to favor stochastic policies and encourage exploration during the training process.

\paragraph{Inverse Reinforcement Learning --}
The goal of IRL is to take as input an MDP sans reward function and a set of demonstration trajectories $\mathcal{D}=\{\tau_1, \tau_2, \cdots, \tau_N\}$ and output a recovered reward function $R$, which denotes the objective trajectories in $\mathcal{D}$ optimize. Our method is based on AIRL~\cite{fu2017learning}, which casts the IRL problem in a generative-adversarial framework. AIRL consists of a discriminator in the form of a reward function and a generator in the form of a learning policy. The discriminator, $D$, is given by $D_\theta(s,a)=\frac{e^{f_\theta(s,a)}}{e^{f_\theta(s,a)}+\pi(a|s)}$, where $f_{\theta}(s,a)$ is the reward function, parameterized by $\theta$, and $\pi(a|s)$ is a policy. The discriminator's objective is to distinguish generated rollouts from demonstrated trajectories via a binary cross entropy, updating the reward function parameter $\theta$ via gradient descent. In other words, the reward function learns to provide higher rewards for the demonstration trajectories than the generated policy rollouts. The policy, $\pi$, is trained to maximize the pseudo-reward function given by $\hat{R}= f_{\theta}(s,a)$, resulting in behaviors close to the demonstrations.

\paragraph{SSRR --}


SSRR learns an idealized reward function from suboptimal demonstrations, consisting of three phases: Phase 1) self-supervised data generation, Phase 2) Noise-Performance characterization, and Phase 3) Reward \& Policy learning. Figure \ref{fig:ssrr_diagram} top depicts SSRR.

\emph{Phase 1)}
For self-supervised data generation, SSRR utilizes the initial reward, $\Tilde{R}$, and the initial policy, $\Tilde{\pi}$, inferred from utilizing AIRL on the suboptimal demonstrations. SSRR chooses random actions uniformly (named ``injecting noise'') to replace the learned policy decision with probability $\eta$ (Equation \ref{eq:noise_injection}) to produce ``noisy'' policies. The trajectories, $\{\tau^i\}\ \forall i$ generated via noise-injection, consist of four elements $\tau^i=\langle\eta^i, \{s_t^i\}, \{a_t^i\}, \{\Tilde{r}_t^i\}\rangle$: the noise parameter, $\eta^i$, a set of states and actions, $\{s_t^i\}$ and $\{a_t^i\}$, and the corresponding initial reward $\tilde{r}_t^i=\Tilde{R}(s_t^i,a_t^i)$. 
\par\nobreak{ \small \noindent
\begin{align}
\label{eq:noise_injection}
    \tau_\eta\sim \pi_\eta(a|s) = \eta U(a) + (1 - \eta)\pi_{\text{AIRL}}(a|s). 
\end{align}}
\emph{Phase 2)}
For noise-performance curve fitting, SSRR empirically characterizes the noise-performance curve. Since SSRR has access to an initial reward estimate for each trajectory, $\Tilde{R}(\tau^i)=\sum_t{\Tilde{R}(s_t^i,a_t^i)}$, SSRR regresses a sigmoid function (see Equation \ref{eq:fit}) towards the cumulative estimated rewards via least squared error minimization. The sigmoid acts as a low-pass filter on the noise-performance relationship from AIRL's initial reward, creating a smooth noise-performance relationship even in the presence of high-frequency neural network output. 
\par\nobreak{ \small \noindent
\begin{align}
\label{eq:fit}
    \sigma(\eta)=\frac{c}{1+\exp(-k(\eta-x_0))}+y_0. 
\end{align}
}
\emph{Phase 3)}
Leveraging the resultant noise-performance curve, SSRR regresses a reward function of trajectory states and actions, as shown in Equation \ref{eq:SSRR_loss}. Here, $R_\theta$ represents parameters of our idealized reward function. After obtaining an accurate reward function, SSRR applies RL to obtain a policy $\pi^*$ that outperforms the suboptimal demonstrations.
\par\nobreak{ \small \parskip0pt \noindent
\begin{align}
\label{eq:SSRR_loss}
    L_{\text{SSRR}}(\theta)=\mathbb{E}_{\tau^i}\left[\left(\left(\sum_{t=0}^T{R_\theta(s_t^i,a_t^i)}\right) - \sigma(\eta^i)\right)^2\right]
\end{align}}
\section{Method}
In this section, we present \textit{systematic} alternatives to self-supervised trajectory generation via noise-injection: 1) Reduction of demonstrations available to the initial imitation learning framework, 2) Reduction of model capacity within the imitation learning framework, and 3) Increase of policy network weight sparsity during the imitation learning.
Relying on a \textit{random} approach such as noise-injection to generate trajectories has two key drawbacks: 1) Performance degradation could be random depending on whether the noise is applied to vital states among the trajectory and 2) Degraded trajectories via noise-injection may not accurately represent the entire scope of suboptimality. 

As discussed in the Preliminaries Section, the SSRR pipeline has three phases consisting of self-supervised data generation, noise-performance curve fitting, and reward $\&$ policy learning. In our extension of SSRR, S3RR, \emph{Phase 1} is modified to utilize different performance-degradation methodologies. In this way, instead of learning one initial policy via AIRL and injecting noise to produce performance-degraded trajectories, we learn several AIRL policies, each representing a different level of degradation. A visualization of the modified Phase 1 procedure (representing all three proposed performance-degradation methodologies) of S3RR is shown in Figure \ref{fig:ssrr_diagram} bottom. We detail the three different degradation methodologies below:

\subsection{Reduction of demonstrations available to AIRL} In our first variant, we alter the number of suboptimal demos available to AIRL in each level of degradation. As machine-learning approaches typically benefit from increased numbers of data, a reduction in the amount of available data should lead to a decrease in performance~\citep{icpram19,althnian2021impact}. As such, we learn several policies, each learning from a varying number of suboptimal demos. Each policy is then used to generate performance-degraded trajectories. The generated degradation dataset is then fed into Phase 2 following the SSRR framework to infer sigmoid parameters to fit the degradation-performance curve.

\subsection{Reduction of Model Capacity within AIRL} In our second variant, we alter the model capacity available to the policy and discriminator networks within AIRL. More specifically, we modify the network sizes to control the model capacities. As highly parameterized networks are typically needed to learn a complete representation of the user's behavior, the performance of machine-learning approaches suffer when the models are under-parameterized~\citep{frankle2018the}. As such, we learn several policies, each varying in the number of parameters for the AIRL networks. Each learned policy can then be used to generate performance-degraded trajectories, which can then be used in Phase 2 following the SSRR framework to infer sigmoid parameters to fit the degradation-performance curve.

\subsection{Increase of Network Weight Sparsity within AIRL} In our third variant, we alter the amount of weight regularization used during training the AIRL policy to control weight sparsity. An increase of L1 regularization (by increasing $\lambda$ in Equation \ref{eq:lambda}) during training will force the policy network to be sparse~\citep{ma2019transformed}, resulting in a lower performing policy.  
As such, we learn several policies, each varying in the amount of regularization used during AIRL. Each learned policy can then be used to generate performance-degraded trajectories, which can then be used in Phase 2 following the original SSRR framework to infer sigmoid parameters to fit the degradation-performance curve. Equation \ref{eq:lambda} shows the policy gradient objective with the added L1 regularization. 
\begin{equation}
    \phi^* = \arg\max \mathbb{E}_{\tau\sim\pi_\phi}[R(\tau) - \lambda ||\phi||_1] 
    \label{eq:lambda}
    \vspace{-5mm}
\end{equation}
\begin{table*}[t]
\caption{Learned Reward Correlation Coefficients with Ground-Truth Reward Comparison between S3RR variants and SSRR. Reported results are Mean $\pm$ Std. Dev. across five trials. Bold results represent the largest correlation  within each domain. }
\begin{center}
\begin{tabular}{P{50pt}P{90pt}P{70pt}P{70pt}P{70pt}P{70pt}}
\hline
\multirow{1}{*}{Domain} & \textit{S3RR} & \textit{S3RR} & \textit{S3RR} & SSRR & SSRR \\
& \textit{Reduction of Demos} & \textit{Model Capacity} & \textit{Weight Sparsity} & Noisy-AIRL & AIRL \\
\hline
\multirow{1}{*}{HalfCheetah-v3} & 0.936 $\pm$ 0.048& 0.977 $\pm$ 0.014& \textbf{0.996 $\pm$ 0.001} & 0.941 $\pm$ 0.025& 0.917 $\pm$ 0.017\\
\hline
\multirow{1}{*}{Ant-v3} & 0.941 $\pm$ 0.021& 0.816 $\pm$ 0.109& 0.741 $\pm$ 0.081& \textbf{0.970 $\pm$ 0.006} & 0.615 $\pm$ 0.024\\
\hline
\label{tab:SSRR_Corr}
\end{tabular}
\end{center}
\end{table*}
\begin{table*}[ht]
\vspace{-20pt}
\caption{Ground-Truth Reward of Policy Inferred via S3RR and SSRR. Best correlation reward functions are chosen to optimize via RL out of the trials of S3RR and SSRR. Bold results represent the highest performing policies in each domain.}
\begin{center}
\begin{tabular}{cccccccc}
\hline
\multirow{2}{*}{Domain} & \multicolumn{2}{c}{Demonstration} & \textit{S3RR} & \textit{S3RR} & \textit{S3RR} & SSRR \\
\cline{2-3} & Average & Best & \textit{Reduction of Demos} & \textit{Model Capacity} & \textit{Weight Sparsity} & Noisy-AIRL \\
\hline
HalfCheetah-v3 & 803.38 & 1246.22 & 572.82 (46\%) & 4968.29 (398.67\%) & \textbf{8021.58 (644\%)} & 2853 (229\%) \\
Ant-v3 & 1297.59 & 1559.44 & 955.94 (61\%) & 943.91 (61\%) & 951.52 (61\%) & \textbf{3944 (253\%)} \\
\hline
\vspace{-20pt}
\label{tab:policy_trained}
\end{tabular}
\end{center}
\end{table*}
\section{Results}
\label{sec:results}
In this section, we show preliminary results with S3RR with two MuJoCo virtual robotic control tasks~\citep{todorov2012mujoco}: HalfCheetah-v3 and Ant-v3, which are common benchmarks in RL and LfD literature.
\begin{figure}[h!]
\begin{center}
  \includegraphics[width=.7\linewidth]{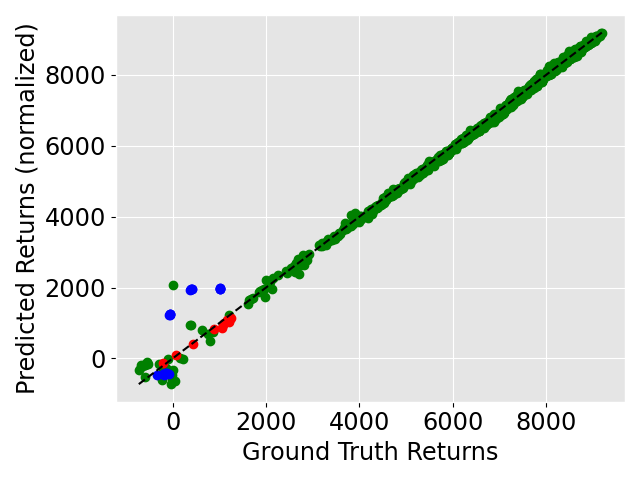}
  \caption{Reward correlation with Ground-Truth reward in HalfCheetah. Red dots represent demonstrations given to AIRL, blue dots represent synthetic degradation data, green dots are unseen “test” trajectories, and the dotted line reflects perfect correlation. S3RR returns are normalized to the range of ground-truth returns.}
  \label{fig:corr_in_half_cheetah}
  \vspace{-15pt}
\end{center}
\end{figure}
We present both the correlation with the ground-truth reward (in Table \ref{tab:SSRR_Corr}) and the performance of the learned policies (in Table \ref{tab:policy_trained}). Each model within S3RR is trained by modifying the respective performance degradation variable. For the reduction of demonstrations available to AIRL, we utilize one, three, six, and ten available demos for training the respective phase 1 policies, with corresponding degradation level $\eta=[1.0, 0.78, 0.44, 0.0]$ by mapping [1,3,6,10] to degradation levels between $1.0$ and $0.0$. For the reduction of network size within AIRL, we vary the number of layers within the AIRL policy and discriminator network to be one, two, three, four, five for training the respective phase 1 policies with corresponding degradation level $\eta=[1.0, 0.75, 0.5, 0.25, 0.0]$. We keep the number of hidden nodes per layer fixed and choose four nodes for HalfCheetah and eight nodes for Ant on each layer. For the increase of network regularization within AIRL, we utilize L1 regularization coefficients of $100, 10, 1, .1, .01$ in training the respective phase 1 policies, with the corresponding degradation level $\eta=[1.0, 0.75, 0.5, 0.25, 0.0]$.

In Table \ref{tab:SSRR_Corr}, we present the learned reward correlation coefficients with the ground-truth reward across the proposed S3RR variants and prior work, SSRR. Specifically, we benchmark against two variants of SSRR, namely SSRR-AIRL and SSRR-Noisy-AIRL, depicting different initial AIRL training methods (See \citet{Chen2020LearningFS} for a detailed comparison between AIRL and Noisy-AIRL). As a comparison, SSRR has 21 noise levels equally spaced between 0 and 1. Across two robotic control domains, HalfCheetah and Ant, we see three variants of S3RR achieve higher correlation compared with SSRR-AIRL, illustrating the benefit of \textit{systematic} self-supervised trajectory generation. Comparing with the previous state-of-the-art SSRR-Noisy-AIRL, S3RR achieves better correlation on HalfCheetah and comparable performance on Ant. We hypothesize S3RR could benefit from the Noisy-AIRL intuition by combining the critic training across AIRL runs with different degradation parameters, which creates a more robust discriminator that can generalize to various state spaces generated by different performing policies. S3RR could also benefit from a recent work on end-to-end SSRR~\cite{aux-airl}. We leave the exploration for future work. We present a depiction of the correlation between S3RR and the ground-truth reward function in HalfCheetah in Figure \ref{fig:corr_in_half_cheetah}. In general, we argue there is likely no one degradation method that fits well~\cite{585893}, let alone the best, to the data in every domain, as observed in our results. As shown in Table \ref{tab:SSRR_Corr}, S3RR Weight Sparsity works the best for HalfCheetah, but is the worst for Ant. We find AIRL policy performs well with regularization coefficients of $1$ or $100$, while with regularization coefficient of $10$, the AIRL policy acts randomly. To that end, our paper proposes a general modification to SSRR that opens the possibility to discover effective degradation processes for broader domains to achieve learning from suboptimal demonstration, despite higher computation needs linear to the number of degradation levels. 

In Table \ref{tab:policy_trained}, we present the performance of the policies learned via S3RR and SSRR. The success of reward learning in S3RR with weight sparsity in HalfCheetah is transferred to the policy, yielding a policy performing $544\%$ better than the best demonstration and $415\%$ better than SSRR-Noisy-AIRL. Besides, S3RR with model capacity also achieves better performance than SSRR-Noisy-AIRL. However, policy learning with S3RR on Ant fails drastically, possibly due to the lower reward correlation compared with SSRR-Noisy-AIRL. In future work, We will conduct further analysis to find the cause of the low policy performance. 

\section{Conclusion}
In this work, we present \textit{Systematic} Self-Supervised Reward Regression, S3RR, to investigate systematic alternatives for trajectory degradation. As alternatives in data generation in imitation learning, we explore 1) a reduction of demonstrations available, 2) a reduction of model learner capacity, and 3) an increase of network weight sparsity. We find that S3RR can achieve higher reward correlation with ground-truth and produce policies performing $544\%$ better than the best suboptimal demonstration and $415\%$ better than the previous state-of-the-art SSRR-Noisy-AIRL.

\section{ Acknowledgments}
We wish to thank our reviewers for their valuable feedback in revising our manuscript. This work was sponsored by MIT Lincoln Laboratory grant 7000437192, the Office of Naval Research under grant N00014-19-1-2076, NASA
Early Career Fellowship grant 80HQTR19NOA01-19ECF-B1, and a gift to the Georgia Tech Foundation from Konica Minolta.

\bibliography{example}  

\end{document}